\pdfoutput=1

\documentclass[11pt]{article}

\usepackage[final]{acl}

\usepackage{times}
\usepackage{latexsym}
\usepackage{booktabs}
\usepackage[T1]{fontenc}

\usepackage[utf8]{inputenc}

\usepackage{microtype}

\usepackage{inconsolata}
\usepackage{amssymb,amsmath,amsthm}

\theoremstyle{definition}
\newtheorem{definition}{Definition}[section]
\usepackage{graphicx}
\graphicspath{{images/}}
\usepackage{listings}
\usepackage{bbm}
\usepackage{amssymb}
\usepackage{amsmath}

\usepackage{listings}
\usepackage{color}

\definecolor{ao(english)}{rgb}{0.0, 0.5, 0.0}
\definecolor{dkgreen}{rgb}{0,0.6,0}
\definecolor{gray}{rgb}{0.5,0.5,0.5}
\definecolor{mauve}{rgb}{0.58,0,0.82}

\definecolor{darkgoldenrod}{rgb}{0.72, 0.53, 0.04}
\definecolor{indianred}{rgb}{0.8, 0.36, 0.36}
\definecolor{mediumseagreen}{rgb}{0.24, 0.7, 0.44}
\definecolor{mediumpurple}{rgb}{0.58, 0.44, 0.86}

\title{A Notion of Complexity for Theory of Mind via\\ Discrete World Models}

\newcommand{\unibo}{1}
\newcommand{\oxford}{2}
\newcommand{\alan}{3}
\newcommand{\leeds}{4}

\author{
   X. Angelo Huang$^{\unibo}$\thanks{First author. Work done while visiting the University of Oxford.} \quad Emanuele La Malfa$^{\oxford, \alan}$ \\
    \textbf{Samuele Marro}$^{\unibo}$ \quad \textbf{Andrea Asperti}$^{\unibo}$ \quad \textbf{Anthony G. Cohn}$^{\alan, \leeds}$ \quad \textbf{Michael Wooldridge}$^{\oxford, \alan}$\\
   $^{\unibo}$DISI, University of Bologna \quad $^{\oxford}$Dept. of Computer Science, University of Oxford \\ 
   $^{\alan}$ The Alan Turing Institute \quad $^{\leeds}$ University of Leeds \\
    \texttt{\href{mailto:xuanqiang.huang@studio.unibo.it}{xuanqiang.huang@studio.unibo.it}} \quad \texttt{\href{mailto:emanuele.lamalfa@cs.ox.ac.uk}{emanuele.lamalfa@cs.ox.ac.uk}} \\
}

\begin{document}
\maketitle
\begin{abstract}
Theory of Mind (ToM) can be used to assess the capabilities of Large Language Models (LLMs) in complex scenarios where social reasoning is required.
While the research community has proposed many ToM benchmarks, their hardness varies greatly, and their complexity is not well defined.
This work proposes a framework inspired by cognitive load theory to measure the complexity of ToM tasks. 
We quantify a problem's complexity as the number of states necessary to solve it correctly. 
Our complexity measure also accounts for spurious states of a ToM problem designed to make it apparently harder. 
We use our method to assess the complexity of five widely adopted ToM benchmarks. 
On top of this framework, we design a prompting technique that augments the information available to a model with a description of how the environment changes with the agents' interactions.
We name this technique Discrete World Models (DWM) and show how it elicits superior performance on ToM tasks.\footnote{Code and data for \textbf{full reproducibility} are available in the Code Material.}

\includegraphics[width=1.25em,height=1.15em]{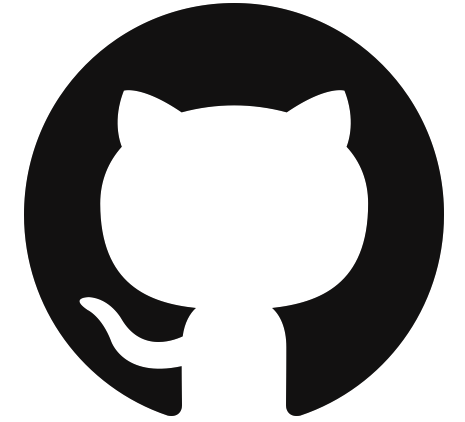}\hspace{.75em}
\parbox{\dimexpr\linewidth-7\fboxsep-7\fboxrule}{\url{https://github.com/flecart/complexity-tom-dwm}}
\vspace{-.5em}
\end{abstract}

\section{Introduction}
\begin{figure}
\centering 
\includegraphics[width=0.48\textwidth]{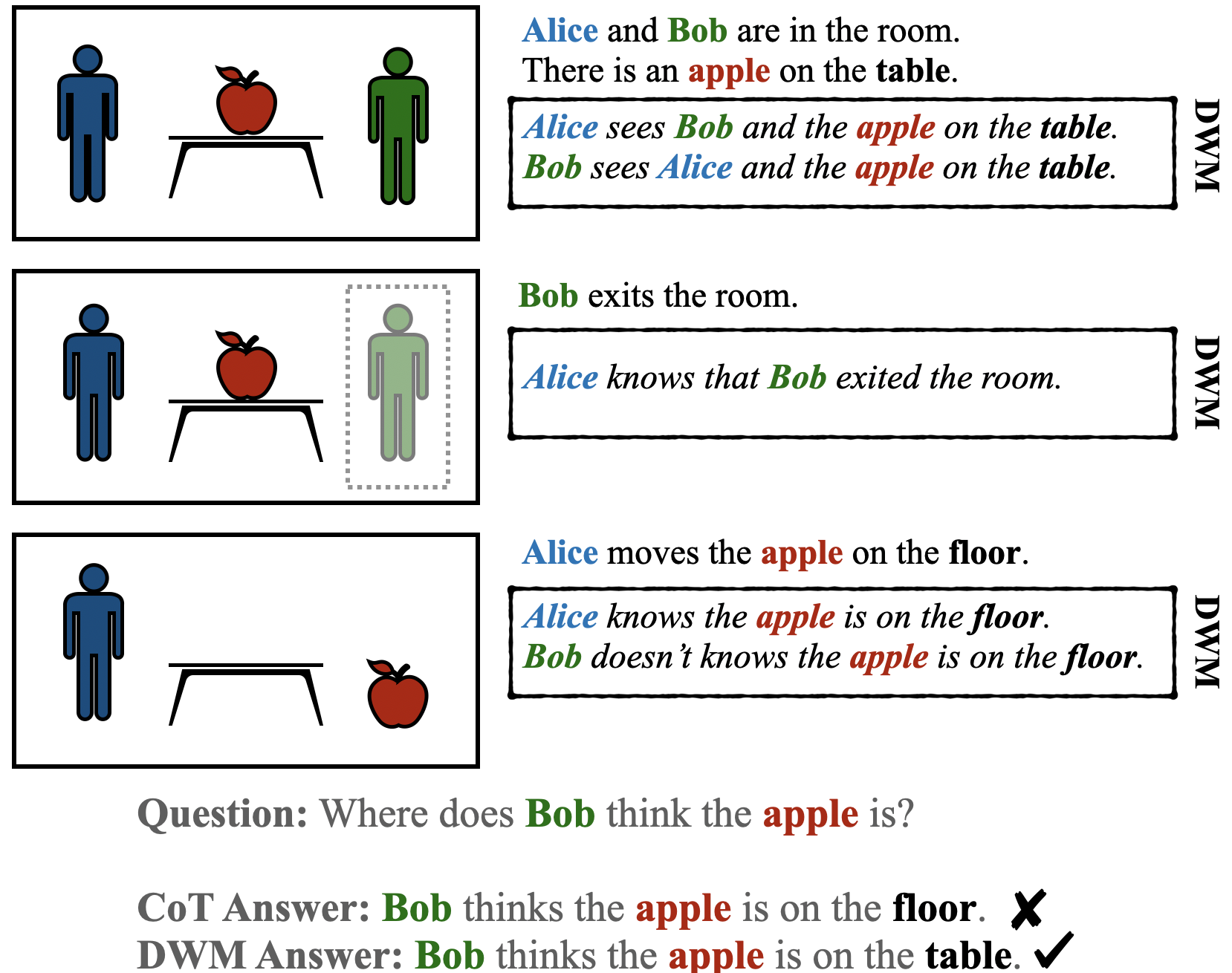} 
\caption{Example of the DWM prompting technique on a classical \textit{Sally-Anne} QA task~\cite{baron-cohenDoesAutisticChild1985}. Inspired by our complexity framework (Section~\ref{sec:complexity}), DWM takes the original task and splits it into sequences, the \textit{state events} (see Def.~\ref{def:state-event}), and prompts the LLMs to describe the states. We show that, in most cases, this aids the LLM in providing correct answers.} 
\label{fig:example} 
\end{figure}
Theory of Mind (ToM) studies how agents form and use beliefs to reason in dynamic environments~\cite{premackDoesChimpanzeeHave1978}.
Originally developed to describe human interactions~\cite{preston2002empathy,tomasello2009cultural} as well as toddlers' psychological development~\cite{wimmerBeliefsBeliefsRepresentation1983,baron-cohenDoesAutisticChild1985}, ToM has been quickly adopted by other fields, including artificial intelligence~\cite{mccarthyAscribingMentalQualities1979, scassellatiTheoryMindHumanoid2002}, bayesian inference~\cite{bakerBayesianTheoryMind2011} and machine learning~\cite{rabinowitzMachineTheoryMind2018}.
In machine learning, ToM has both descriptive and prescriptive usage: on the one hand, ToM benchmarks assess the capabilities of a model in complex environments; on the other, ToM's frameworks such as \emph{theory-theory}~\cite{gopnikTheoryTheory1994} and \emph{simulation theory}~\cite{churchland2013folk} have been widely adopted to test the proficiency of Large Language Models (LLMs) in social tasks where humans excel~\cite{strachan2024testing}.

In this work, we propose a framework to characterise a ToM benchmark's difficulty, i.e., its \textbf{complexity}, as the number of \emph{state events} that are sufficient to track the state of an object, including $k^{\text{th}}$-order beliefs motivated by theoretical parallelisms with Sweller's cognitive load theory~\cite{swellerElementInteractivityIntrinsic2010}.

We characterise the complexity of five standard ToM benchmarks, from false belief to commonsense and social reasoning, and compute their complexity as a proxy of their inherent difficulty. 
Inspired by prompting techniques that split a task into elementary sub-problems that are solved sequentially, like Tree of Thoughts~\cite{yaoTreeThoughtsDeliberate2023} and least-to-most prompting~\cite{zhou2023leasttomost}, we introduce a technique that stimulates a model's reasoning capabilities via Discrete World Models (DWM). DWM leverages the notion of statefulness via a succinct and coherent representation of each \emph{state events}, as illustrated in Figure~\ref{fig:example}.
We test DWM on ToMi~\cite{leRevisitingEvaluationTheory2019}, MindGames~\cite{sileoMindGamesTargetingTheory2023}, Adv-CSFB ~\cite{shapiraCleverHansNeural2023}, SocialIQA~\cite{sapSocialIQaCommonsense2019}, and FANToM~\cite{kimFANToMBenchmarkStresstesting2023}, eliciting superior performance than Chain of Thoughts (CoT)~\cite{weiChainofThoughtPromptingElicits2023} and Tree of Thoughts (ToT)~\cite{yaoTreeThoughtsDeliberate2023} on those problems whose \emph{state spaces} are informative.
We further assess whether memorisation affects a model's performance, and we discover that while this phenomenon happens for standard benchmarks such as ToMi~\cite{leRevisitingEvaluationTheory2019}, with input-output pairs that can be retrieved \emph{word for word} via prompting, it does not strongly correlate with a drop of performance on memorised ToM benchmarks.
We conduct our experiments on a variety of open- and closed-source LLMs, including GPT-3.5-Turbo, GPT-4~\cite{OpenAI2023GPT4TR}, LLaMA3-70B~\cite{llama3modelcard,dubey2024llama} and Mixtral~8x7B~\cite{jiang2024mixtral}.
In summary, in this paper: 
\begin{itemize}
    \item We introduce the concept of \textbf{complexity} of a ToM task to quantify the hardness of keeping track of the elements (e.g., agents' beliefs or objects' states) that are sufficient to produce the correct answer to different problems inspired by frameworks in cognitive science.
    \item We propose DWM, a simple yet effective prompting technique that improves a model's capability by making \textbf{implicit} information explicit while not necessitating \emph{exogenous information} (i.e., it does not require RAG or fine-tuning).
\end{itemize}
We consider our work a step towards a framework that formalizes the hardness of a ToM problem in an unambiguous way, inspired by the theory of World Models~\cite{wongWordModelsWorld2023}.







\section{Related Work}

Over 40 years of research on ToM in psychology~\cite{premackDoesChimpanzeeHave1978, baron-cohenDoesAutisticChild1985, dennettIntentionalStanceTheory1988, wellmanDevelopmentTheoryMind2017} on human development has created a fertile ground for the development of these ideas in adjacent fields. In the last decade, many works studied ToM in artificial intelligence and machine learning~\cite{bakerBayesianTheoryMind2011, rabinowitzMachineTheoryMind2018}, with applications to multi-agent systems and reinforcement learning~\cite{gronauer2022multi}.
More recently, the rise in popularity of LLMs shifted the interest towards understanding and benchmarking large models' capacity to solve increasingly complex ToM tasks~\cite{aru2023mind,zhou2023far,mahowald2024dissociating}. 
While some researchers believe LLMs have already become proficient in solving ToM tasks~\cite{bubeckSparksArtificialGeneral2023, kosinski2023theory, strachan2024testing}, others show scepticism and illustrate cases where they fail on trivial variations of well-known problems~\cite{ullmanLargeLanguageModels2023,shapiraCleverHansNeural2023, sapNeuralTheoryofMindLimits2023}.
In a joint effort between computer scientists and psychologists, many ToM benchmarks have been developed and used to test neural-network models, including LLMs~\cite{gandhiBabyIntuitionsBenchmark2022,chen2024tombench,strachan2024testing}.
Recently, concepts such as World Models~\cite{haWorldModels2018} have found applicability 
mostly as discrete prompting techniques in conjunction with optimisation procedures~\cite{hao-etal-2023-reasoning,moghaddam2023boosting}.
Researchers have found evidence of an emergent internal representation (e.g., World Model's surrogates) of the state games~\cite{li2022emergent, toshniwal2021learningchess} and state-tracking abilities~\cite{li2021implicit,kim2023entity,kim2024codepretraining}, necessary for correct belief tracking in ToM problems.
Cognitive load theory emerged in the late eighties with Sweller's work on human problem solving~\cite{swellerCognitiveLoadProblem1988}. Most measures of cognitive load are based on subjective reports from humans~\cite{swellerMeasuringCognitiveLoad2011}. Even though some attempts at automatic cognitive load measures have been present~\cite{yinAutomaticCognitiveLoad2007}, they have not been widely adopted in the community.
The works that are more similar to our complexity framework are only tangentially related to ToM. 
Inspired by the work in~\cite{zhou2023leasttomost} and the results in~\cite{zhou2023far}, our prompting technique is inspired by~\cite{park2023generative} and~\cite{nye2021show}: the former develops an architecture to record the agent’s experiences. The latter proposes a prompting technique that forces a model to express the intermediate computational steps to solve a problem.

\begin{figure*}
\centering 
\includegraphics[width=1\textwidth]{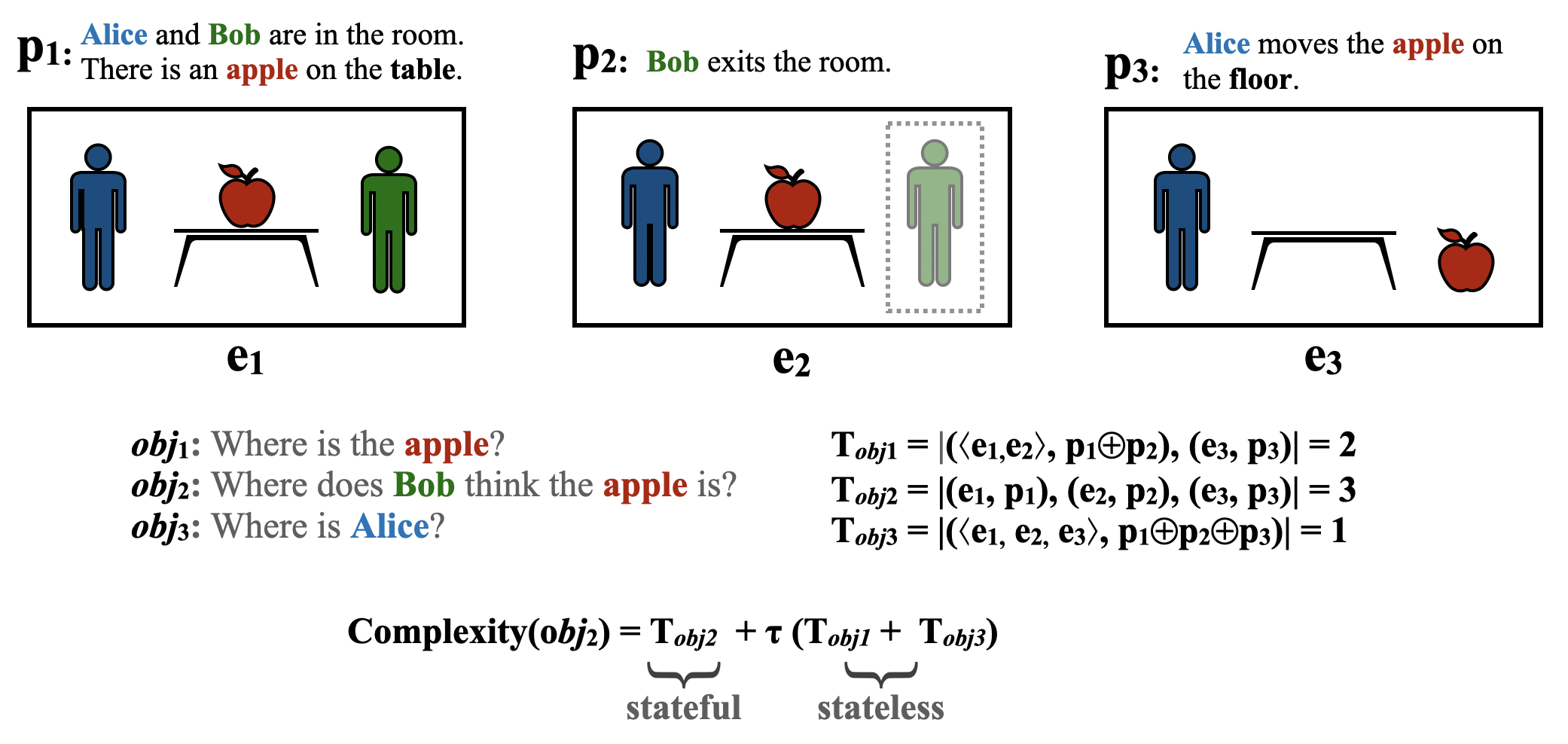} 
\caption{
How statefulness and statelessness (Def.~\ref{def:partition}) are computed for the motivating example in Fig.~\ref{fig:example}. For ${obj_1}$, an optimal split to track the \textbf{\textcolor{purple}{apple}} merges the first two states and chunks of the input prompt. For ${obj_2}$, which involves the $1^{\text{st}}$-order belief of \textbf{\textcolor{ao(english)}{Bob}}, the statefulness is higher, with $e_2$ that cannot be merged with $e_3$ as it introduces partial observability. The complexity of the task (bottom) is computed as per Eq.~\ref{def:tom-complexity}, 
where the complexity of objects that are not directly relevant to the question/answer is discounted.
}
\label{fig:statefulness} 
\end{figure*}

\section{Methodology} 
In this section, we introduce a notion of complexity for ToM problems
which 
quantifies the hardness of a problem as the number of \emph{computational steps} humans take to solve them and compare it with Sweller's cognitive load theory.
We then present the DWM prompting technique within the complexity framework and show how it differs from standard methods like CoT and ToT. 
We further characterise its efficiency with the number of input/output tokens and queries to a model as the control variables.

\subsection{On the Complexity of ToM}\label{sec:complexity}

The need to provide
a consistent representation of the environment, including each agent's beliefs, inspired us to characterise the complexity of a ToM problem in terms of \textbf{sufficient elements to track} to output the correct result.
Consider a problem prompt $p$, expressed in natural language, that describes how multiple agents interact with an environment object $\mathbf{obj}$, as illustrated in Figure~\ref{fig:statefulness} (top). In our framework, an object can be the state of the apple as well as the $k^{\text{th}}$-order belief of an agent about the apple position. Our framework naturally extends to multiple objects by considering their union.

Suppose that in $p$, the state of $\mathbf{obj}$ is modified $T>0$ times, thus identifying $T$ unique configurations, namely $E_{\mathbf{obj}}=\{e_1, .., e_T\}$.
To correctly solve a ToM task where $p$ is complemented by a query about $\mathbf{obj}$, a model should distinguish between the interactions that modify the configuration of $\mathbf{obj}$, i.e., the \textbf{stateful} states, from those that modify any other \textbf{stateless} object $Obj \setminus \mathbf{obj}$, i.e., those that  one does not need to track.

We first show how to define the cost of tracking a task's {stateful} states, which we complement with that of the {stateless}. Both definitions concur in defining the \textbf{complexity} of a ToM task.

\subsubsection{Stateful and Stateless Complexity}
For a ToM task, expressed as $p$, that describes the evolution of an environment where an unknown number of atomic iterations $T$ modifies $\mathbf{obj}$ or its perception, each environment state $e_t \in E_{\mathbf{obj}}$ can be coupled with the prompt prefix $p_{\le t} \text{ s.t. } p_{\le t} \oplus p_{> t} = p$, that describes such configuration.
We denote $(e_t, p_{\le t})$ as a generic \emph{state description}, as illustrated in Figure~\ref{fig:statefulness} (top).

\begin{definition}[State event]\label{def:state-event}
A \emph{state event} for an object $\mathbf{obj}$
is an event that links 
adjacent
\emph{state descriptions} that involve, for both the environment state $e_t$ and the sub-prompt $p_{\le t}$, a state change of $\mathbf{obj}$. Formally, we define a relation, $F_\mathbf{obj}$, to specify which pairs of state descriptions form a state event:
$F_\mathbf{obj}((e_t,p_{\le t}),(e_{t+1},p_{\le t+1})) \equiv  
e_t \neq e_{t+1} \ \wedge \ p_{\le t+1} =  p_{\le t} \oplus p_{t+1} $
    where $1 \leq t  \le \lvert p\rvert.$  ($\lvert p \rvert$ denotes the number of atomic prompts.) and $\oplus$ is the string concatenation operator.
\end{definition}
Thus a \emph{state event} $F_{\mathbf{obj}}$ identifies those \emph{state descriptions} $(e_t, p_{\le t})$ which have a successor $(e_{t+1}, p_{\le t+1})$ where $\mathbf{obj}$ has changed its configuration.



In the context of ToM tasks, a \textit{state event} could be a person who moves an object, exits (thus introducing partial observability) or witnesses a change in the environment (as now the description of the environment will take that change into account), as illustrated Figure~\ref{fig:statefulness}  (middle).
Our prompting technique, namely DWM (Section~\ref{sec:dwm}), aims at making implicit observations about objects explicit.

We finally introduce the notion of \emph{partition function} to connect the \textbf{maximum number} of non-empty \emph{state events} relative to a prompt. 
Such a notion will serve as the building block to compute the complexity of a ToM problem.

\begin{definition}[Partitions]\label{def:partition}
A \emph{partition } $part_\mathbf{obj}$  w.r.t. $\textbf{obj}$ identifies those \emph{ state events} which partition a ToM prompt $p$ into sequential segments where $\mathbf{obj}$ changes its value.
Formally:

\begin{equation}
\begin{aligned}
\text{Let } \ part_\mathbf{obj} = \{(e_t,p_{\le t}): \\
F_\mathbf{obj}((e_t,p_{\le t}),(e_{t+1},p_{\le t+1})) \\
\, \wedge \, e_t \in E_\mathbf{obj}\} 
\end{aligned}
\end{equation}
\end{definition}
Def.~\ref{def:partition} describes an optimal partition, $part_\mathbf{obj}$ of \emph{state descriptions} that 
covers all the relevant changes to $\mathbf{obj}$. The partition is represented by the set of event descriptions where $\mathbf{obj}$ changes its description immediately after. Note that this set of event descriptions is unique for any $\mathbf{obj}$. 

\subsubsection{The Complexity of a ToM Task}
We can now define the notion of \textbf{statefulness} of a ToM task specified as a prompt $p$ as the size of Eq.~\ref{def:partition}, namely $T_{\mathbf{obj}} = \lvert part_{\mathbf{obj}} \rvert$. The process of computing the statefulness of an object or its belief is illustrated in Fig.~\ref{fig:statefulness}.

For a ToM task where the question to solve relates to an object $\mathbf{obj}$, one must ensure that changes to any other object, namely $Obj \setminus \mathbf{obj}$, do not affect $\mathbf{obj}$. 
While tracking the evolution of what is irrelevant to answer the question is unnecessary, a computation model must assess whether a particular environmental change affected $\mathbf{obj}$.
We thus introduce the notion of \textbf{statelessness}, i.e., the cost of discerning whether a change in the environment affects $\mathbf{obj}$.
The computation is similar to that of Def.~\ref{def:partition},
except that $\mathbf{obj}$  is replaced with
any object in $Obj \setminus \mathbf{obj}$; 
however
for stateless objects, we introduce a discount factor $\tau$ to penalise the complexity of \emph{state events} that do not affect $\mathbf{obj}$.
Mathematically, we formalise the statelessness of a ToM task involving an object $\mathbf{obj}$ as $\tau \sum_{obj \in Obj \setminus \mathbf{obj}} T_{obj}$.

Finally, we formalise the complexity of a ToM task w.r.t. an object $\mathbf{obj}$ as the complexity of the stateful states plus the (discounted) sum of the others (i.e., stateless).
Namely:
\begin{equation}\label{def:tom-complexity}
\begin{aligned}
T_{\mathbf{obj}} + \tau \sum_{obj \in Obj \setminus \mathbf{obj}} T_{{obj}}
\end{aligned}
\end{equation}

The process of computing the complexity of a ToM task is illustrated in Figure~\ref{fig:statefulness}.






    \begin{figure*}
    \centering 
    \includegraphics[width=1.0\textwidth]{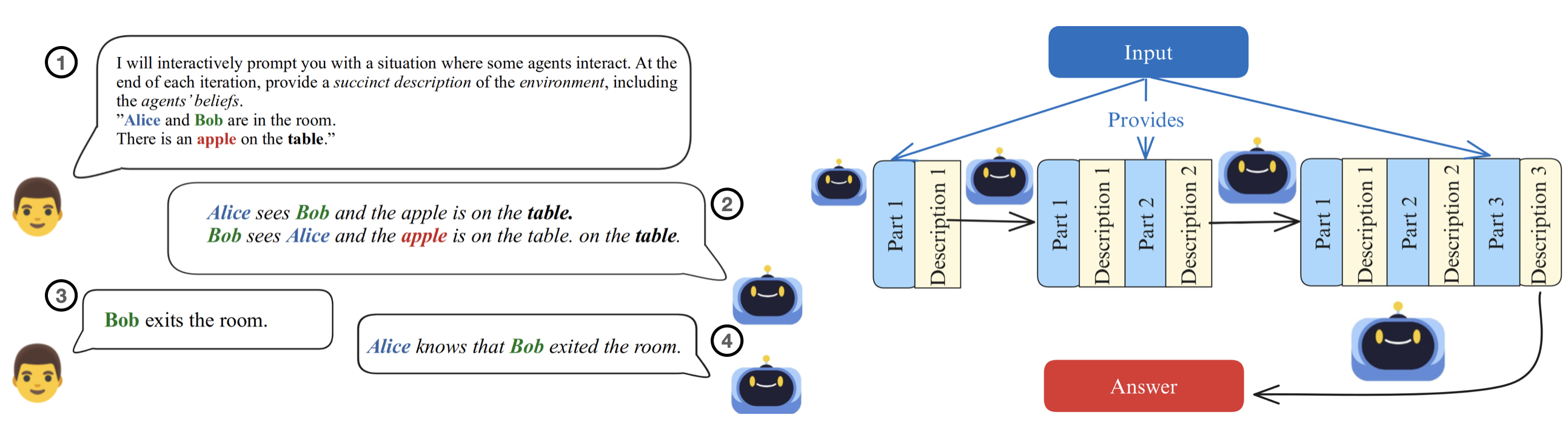} 
\caption{Left: illustration of DWM prompting as per the example in Figure~\ref{fig:example}. We interactively prompt an LLM with a ToM problem, asking to provide a succinct representation of each agent's beliefs. Right: schematic presentation of the DWM method. We first break the input string into $T$ \emph{state descriptions}. Then, for each part, we ask the LLM to provide the \textbf{state event} of the environment and how it changes. In the last step, every part of the input and description is fed to the LLM with another prompt to get the answer for the task.} 
    \label{fig:dwm} 
    \end{figure*}

\subsubsection{A parallelism from the Cognitive sciences}
Understanding how humans solve complex problems has long served as a valuable source of inspiration for advancing machine intelligence. Thought frameworks in the cognitive sciences, such as Kahneman's Dual-process theory~\cite{kahnemanThinkingFastSlow2011}, have greatly influenced various fields, including artificial intelligence.
In this work, we draw ideas from another theory, less known in the community, the cognitive load theory~\cite{swellerCognitiveLoadTheory1994, swellerCognitiveLoadProblem1988, swellerElementInteractivityIntrinsic2010}.
According to this theory, three main factors influence the mental effort humans exert when solving a particular task or learning new information: intrinsic, extraneous and germane load.
The \textbf{intrinsic load} measures the natural difficulty of a certain task, the information that \textit{needs} to be digested before answering the question. This relates to the complexity of the material itself.
The \textbf{extraneous load} concerns the presentation of the information in the problem. For example, if questions phrased in a complex manner are used with a child, it would be much more difficult to understand and answer correctly compared to easier phrasing. Similarly, if many confounding sentences or sentences that do not matter in answering a question are present in the text, we expect an LLM to be worse, suggesting a weak similarity between the means of reason of these models and humans.
Finally, \textbf{germane load} estimates the working memory resources needed to understand the important parts of the problem, i.e. the intrinsic load. If part of the memory is devoted to the extraneous load, then the germane load is diminished, suggesting a positive correlation with intrinsic load and a negative correlation with extraneous load.

\subsubsection{A comparison with the Cognitive Load Theory}


Our framework, summarized in figure~\ref{fig:statefulness}, has two main parts: stateful and stateless complexity. These notions have some similarities with, respectively, the \textit{intrinsic load} and \textit{extraneous load}.
Stateful complexity provides a measure on the sentences that are needed to answer the question correctly and must be adequately represented in memory. In a similar manner, intrinsic load concerns on the needed information to correctly analyze a task. 
Likewise
stateless complexity yields information about the confounding or irrelevant sentences and phrases in the text akin to extraneous load.
In our setting, \textit{germane load} could be interpreted as the ratio of the stateful and stateless complexity: higher ratio means higher density of useful sentences in answering a question. This notion of load could be used as a basis of an objective measure on the quality of a question-answering sample: given the same quantity of cognitive load, i.e. complexity, we would like to have a simple presentation with correct information, maximizing the germane load. If the cognitive load hypothesis applies to LLMs, maximizing the germane load would lower the complexity of the tasks given to a model, and therefore it would aid the model to answer questions more accurately.

\subsection{Discrete World Models}
We first introduce the background notation for prompting LLMs and assessing their accuracy on a standard classification task. We then propose our technique, namely DWM, which we eventually connect with the notion of statefulness of a ToM task.
\paragraph{Background notation.}
A (Large) Language Model is a function that predicts the next token (out of a finite vocabulary) conditioned on the sequence of previously fed/generated tokens, namely $\psi: \mathbf{v} \in V^* \xrightarrow{} v \in V$. Such a mechanism can be used to sample multiple token outputs until an \texttt{`end-of-text'} token is predicted, by invoking $\psi$ in an auto-regressive fashion, i.e., $\psi(v | \mathbf{v})$.
In our setting, a problem is specified as a tuple $(p, Q)$, where $p$ is a ToM problem and $Q$ is a \emph{query} function that modifies $p$ according to a prompting technique, namely $Q: p \xrightarrow{} p'$.
The LLM's output $y$ for an input $Q(p)$ is then compared for correctness against an oracle $\Omega$, i.e., $\Omega: \psi(Q(p)) \xrightarrow{} \{0, 1\} $, where $1$ means correct classification ($0$, otherwise).
On a sample of $N>0$ ToM problems, the accuracy of a model $\psi$ is then measured as $\frac{1}{N}\sum_{i=1}^{N}\Omega(\psi(Q(p_i))$, i.e., the average number of times a model is correct in its prediction.

\subsubsection{Discrete World Models via Prompting}\label{sec:dwm}
Given a ToM problem $p$ and a constant $T \le |p|$, where $|p|$ is ideally measured as the number of state changes in the problem,
we can rewrite $p$ as $p_1 \oplus p_2 \oplus \cdots \oplus p_T$. Our \emph{query} function adds a standard preamble $x$ similar to that of CoT. 
DWM inserts, after each ``split'' $p_t$, an additional prompt $w$ like \texttt{`Now, provide a succinct description of the state of the environment and each agent's belief.'} and query an LLM to provide a representation of the current \emph{state description} of the environment.
An LLM is initially queried with $x \oplus p_1 \oplus w$, and the answer $a_1$ is concatenated to the next query, i.e., $\psi(x \oplus p_1 \oplus w \oplus a_1 \oplus p_2 \oplus w)$ to retrieve $a_2$ 
. The process is carried on for each of the $T$ chunks, and, at the end, $y$ is concatenated to eventually prompt the model for the correct answer to $p$.

Let $a_{1} = \psi(x \oplus p_{1} \oplus w)$,
$a_{t} = \psi(x \oplus p_{1} \oplus w \oplus a_{1} \oplus p_{2} \oplus\dots \oplus a_{t-1} \oplus p_{t} ) = \psi(x \oplus \left(\bigoplus_{i=1}^{t-1} p_i \oplus w  \oplus a_i  \right) \oplus p_{t})$, then, the final query is
\begin{equation}\label{eq3}
    \psi(x \oplus \left(\bigoplus_{t=1}^{T} p_t \oplus w \oplus a_t \right ) \oplus y)
\end{equation}

In this sense, our \textbf{partition function} (Def.~\ref{def:partition}) consists of splitting a prompt into sequential chunks of the prompt, while the LLM is prompted to provide each \emph{state event} at time $1 \le t < T$ as $e_t = \psi(x \oplus \left(\bigoplus_{i=1}^t p_{i} \oplus w \oplus a_{i} \right ) \oplus \omega)$.
The process of prompting a model with DWM is illustrated in Figure~\ref{fig:dwm}.

\subsubsection{On the Complexity of DWM}
DWM progressively calls an LLM $T>0$ times to generate informative states. For a ToM problem of length $n$ (i.e., the number of input tokens), which we assume, w.l.o.g., that can be split into $k$ chunks of approximately the same length $\lvert x \oplus p_i \oplus w \rvert  = \frac{n}{T}$, the number of tokens generated by an LLM is 
of the order of $\mathcal{O}(\sum_{t=1}^{T}\lvert x \oplus \left(\bigoplus_{i=1}^{t-1} p_i \oplus w \oplus a_i \right)  \rvert)$, where $p_t$ ($a_t$) is the portion of the problem (answer) prompted (retrieved) at iteration $t$. With the further assumption that each answer retrieved at split $t\le T$ has the same length $o$, the complexity is further simplified to be asymptotic to $\mathcal{O}((\frac{n}{T}+o)^2)$.
Compared to CoT, whose complexity is $\mathcal{O}(n + o)$, DWM requires an additional linear number of calls to the model.
On the other hand, ToT with the same number of splits $\frac{n}{T}$ and $m>1$ experts results in even higher complexity, i.e., asymptotic to $\mathcal{O}(m(\frac{n}{T}+o)^2)$.

\section{Experimental Evaluation}
\begin{figure*}
\includegraphics[width=0.998\textwidth]{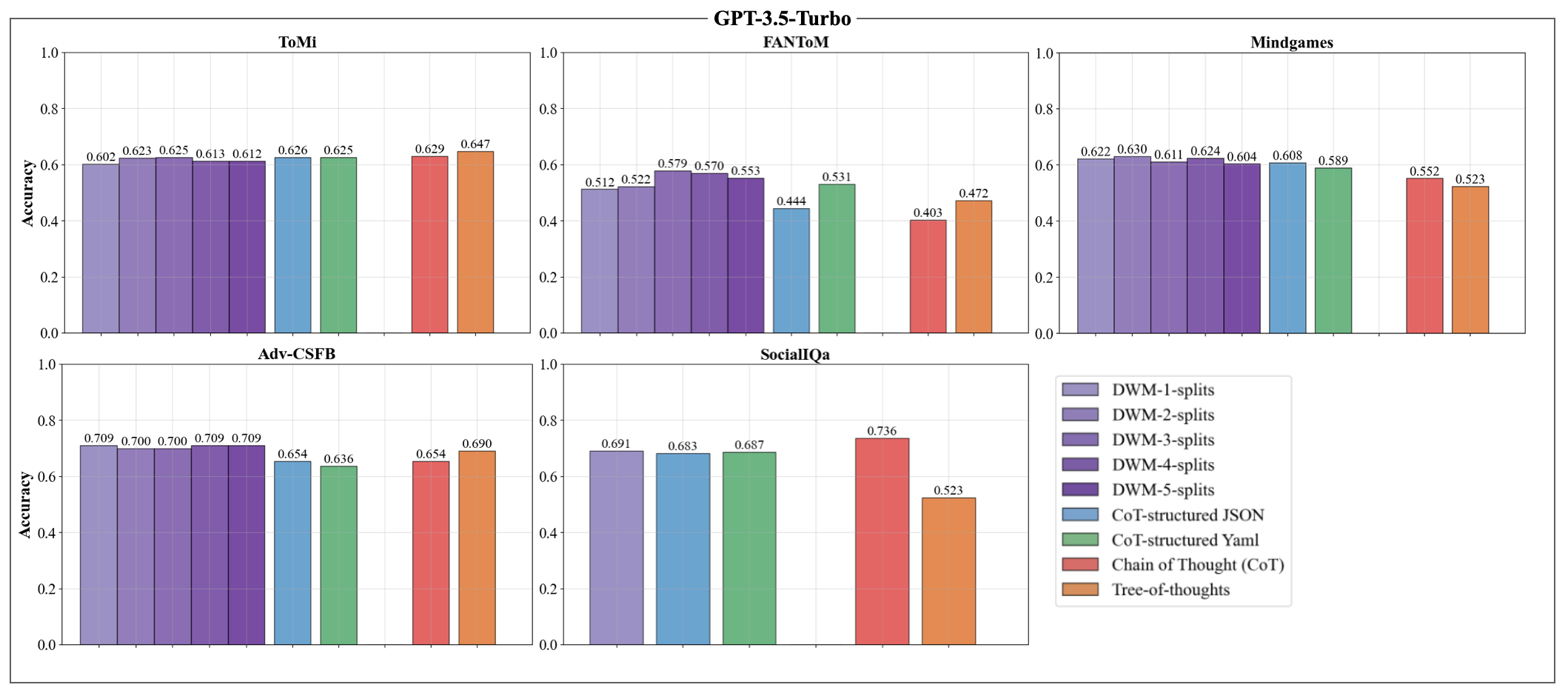} 
\includegraphics[width=1\textwidth]{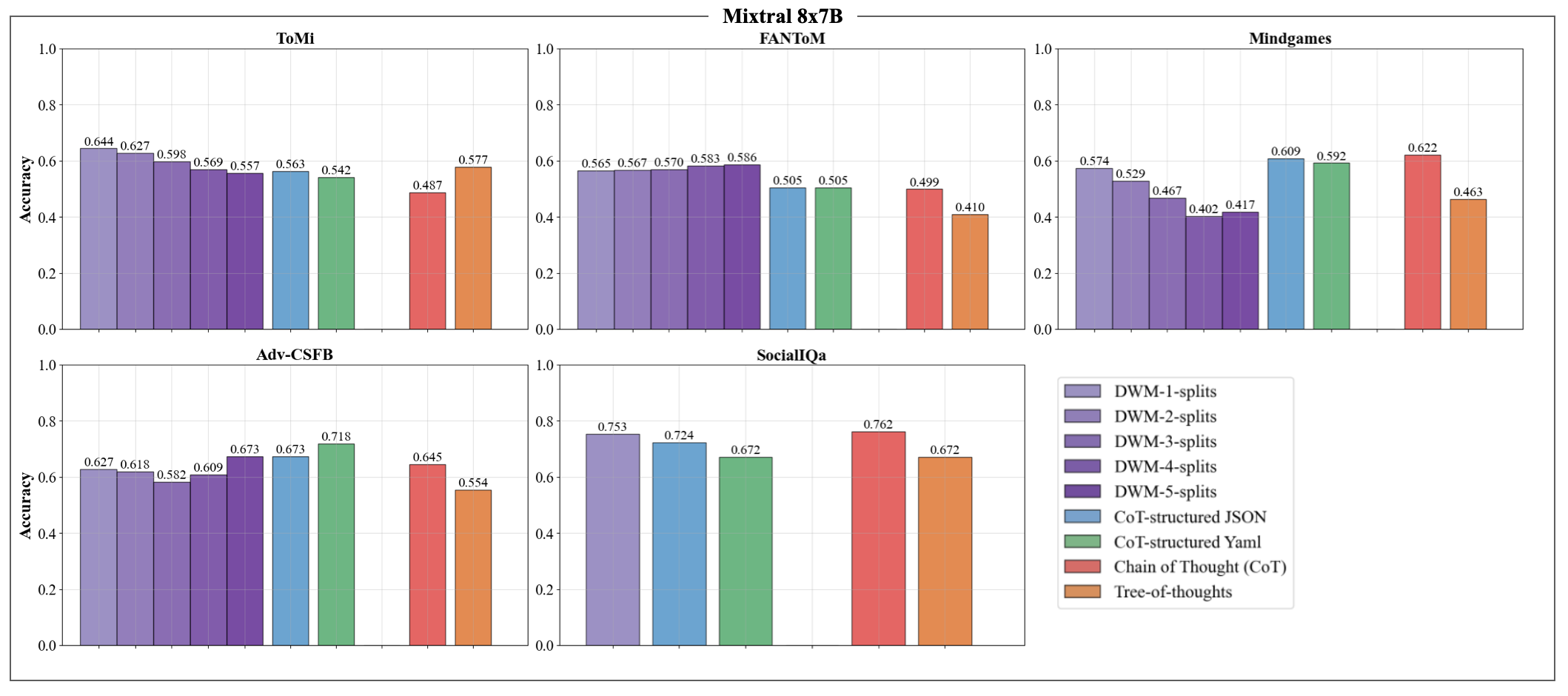} 
\caption{Benchmarks of GPT-3.5-Turbo (top) and Mixtral 8x7B (bottom) models on different ToM tasks for DWM (one to five splits), CoT, ToT and structured prompts (JSON and Yaml).}\label{fig:gpt-3.5-dwm-results}
\end{figure*}
The experiments are organised as follows. We first test the performance of DWM on ToMi~\cite{leRevisitingEvaluationTheory2019}, MindGames~\cite{sileoMindGamesTargetingTheory2023}, Adv-CSFB ~\cite{shapiraCleverHansNeural2023}, SocialIQA~\cite{sapSocialIQaCommonsense2019}, and FANToM~\cite{kimFANToMBenchmarkStresstesting2023}, comparing it with CoT~\cite{weiChainofThoughtPromptingElicits2023}, ToT~\cite{yaoTreeThoughtsDeliberate2023} and prompting with structured data (struct), i.e., the model is queried to first represent the problem in a structured format such as JSON or Yaml.
We further show that ToMi has been memorised \emph{word for word} by GPT models, with CoT (and any technique that leaves the input unchanged) being the best-performing method.
We then quantify the complexity of the benchmarks introduced above and highlight the correlation with the models' performances. Our framework shows complexity ranges between easy and hard problems, even within a benchmark.  
We conduct our experiments on GPT-3.5-Turbo, GPT-4~\cite{OpenAI2023GPT4TR}, LLaMA3-70B~\cite{llama3modelcard,dubey2024llama} and Mixtral~8x7B~\cite{jiang2024mixtral}.

\begin{table*}
\centering

  \resizebox{\textwidth}{!}{%
\begin{tabular}{|c | c c c c c|}
 \hline
    & \textbf{ToMi} & \textbf{FANToM} & \textbf{Mindgames} & \textbf{Adv-CSFB} & \textbf{SocialIQa}  \\
 \hline

    \textbf{Memorisation - perfect match}  & 52\% & 35\% & 2\% & 0\% & 0\% \\ 
 \hline
 
    \textbf{Memorisation - fuzzy}  & $89 \pm 15$\% & $74 \pm 24$\% & $64 \pm 18$\% &  $51 \pm 11$\% & $40 \pm 12$\% \\ 
 \hline
 \hline
 
    \textbf{DWM}  & 0.625 & \textbf{0.579} & \textbf{0.618} &  \textbf{0.8364} & 0.691  \\
 \hline

    \textbf{CoT} & \textbf{0.629} & 0.403  & 0.552 & 0.7091 & \textbf{0.736}     \\
 \hline
 
\end{tabular}
}
\caption{Summary of the memorisation test on five ToM benchmarks. We prompted GPT-3.5-Instruct to predict the continuation of $100$ randomly sampled test points. We computed the exact and fuzzy memorisation rate (second row, similarity score computed via the Levenshtein distance, see the \href{https://github.com/seatgeek/thefuzz}{thefuzz} package), which we complement with the best performance across models of CoT and DWM.}
\label{tab:memorisation}
\end{table*}

\begin{table*}
  \centering

  \resizebox{\textwidth}{!}{%
\begin{tabular}{|c | c c c c c|}
 \hline
    & \textbf{ToMi} & \textbf{FANToM} & \textbf{Mindgames} & \textbf{Adv-CSFB} & \textbf{SocialIQa}  \\
 \hline

    \textbf{Statefulness}  & $2.62 \pm 1.68$ & $2.44 \pm 0.96$ & $1.22\pm 0.90$ & $3.24 \pm 1.35$ & $1. \pm 0.$ \\ 
 \hline
 
    \textbf{Statelessness}  & $ 4.27 \pm 2.1$ & $59.42 \pm 18.91$ & $5.24 \pm 2.71$ &  $2.86 \pm 1.34$ & $1.14 \pm 0.447$ \\ 
 \hline
 \hline

    \textbf{DWM - Best Split}  & 3 & 3 & 1 &  4 & 1  \\
 \hline
 
\end{tabular}
}
\caption{Summary of the statefulness and statelessness of different ToM benchmarks. At the bottom, the value of the split that guarantees max performance of GPT-3.5-Turbo with DWM, which we notice is strongly correlated with the statefulness of each benchmark.}
\label{tab:statefulness}
\end{table*}

\subsection{DWM on ToM Benchmarks}
\begin{figure*}
\centering 
\includegraphics[width=1\textwidth]{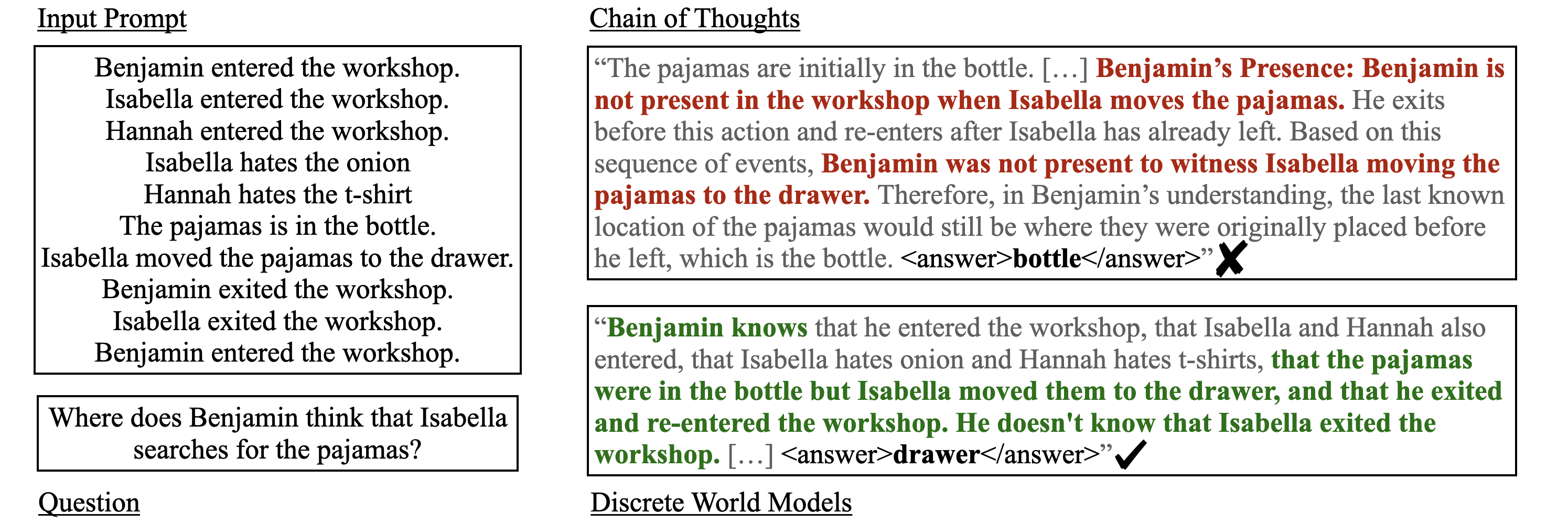} 
\caption{Example of a real ToMi example where GPT-4 fails when prompted with CoT, yet succeeds with DWM. CoT elicits an untruthful reasoning process (in \textbf{\textcolor{purple}{red}}), while DWM correctly informs the model with the implicit information about Benjamin's first-order belief (in \textbf{\textcolor{ao(english)}{green}}). More examples are reported in the Appendix, Section~\ref{a:dwm-vs-cot}.} 
\label{fig:dwm-explicit} 
\end{figure*}
We report results for GPT-3.5-Turbo and Mixtral~8x7B on the five ToM benchmarks: for reasons of space, results for LLaMA3-8B, LLaMA3-70B and GPT-4 are reported in the Appendix, Section~\ref{a:dwm-results}.
As illustrated in Figure~\ref{fig:gpt-3.5-dwm-results} (top), DWM improves the performance of GPT-3.5-Turbo on Mindgames, FANToM and Adv-CSFB by a solid margin.
On SocialIQa, which has very short inputs, DWM performs slightly worse than CoT but better than ToT. On the other hand, on ToMi, the best prompting techniques are CoT and ToT. 
While one may think memorisation plays a role in boosting the performance of LLMs with these prompting techniques, in the next section, we provide evidence this hypothesis is not necessarily true.
With Mixtral~8x7B (Fig.~\ref{fig:gpt-3.5-dwm-results} (bottom)), DWM improves the performance on ADVcsfb, FANToM, ToMi and Mindgames, and 
reaches
that of CoT on SocialIQa.

\paragraph{DWM elicits more informed \emph{state spaces}.}
We qualitatively analysed the information elicited by an LLM when prompted with DWM and discovered that it forces a model to output information \textbf{not explicitly} available in the prompt.
Consider the ToMi example in Figure~\ref{fig:dwm-explicit} where GPT-4 is prompted with a situation where agents interact and are then queried with the first-order belief of Benjamin.
With CoT, the model makes an erroneous assumption about the presence of Benjamin and Isabella in the room.
On the other hand, when prompted with DWM, GPT-4 provides an informative description of each \emph{state space}, particularly the knowledge and the uncertainty of each agent's beliefs, and eventually answers correctly.
One example per benchmark is available in the Appendix, Section~\ref{a:dwm-vs-cot}, while many more are available for inspection in the Code Supplementary Material. 
This
phenomenon is ubiquitous to all the ToM tasks we tested, a hint that DWM 
may elicit
the ToM capabilities of LLMs without requiring external information or solvers.

\paragraph{Memorisation in Theory of Mind.}
\begin{figure}
\centering 
\includegraphics[width=0.5\textwidth]{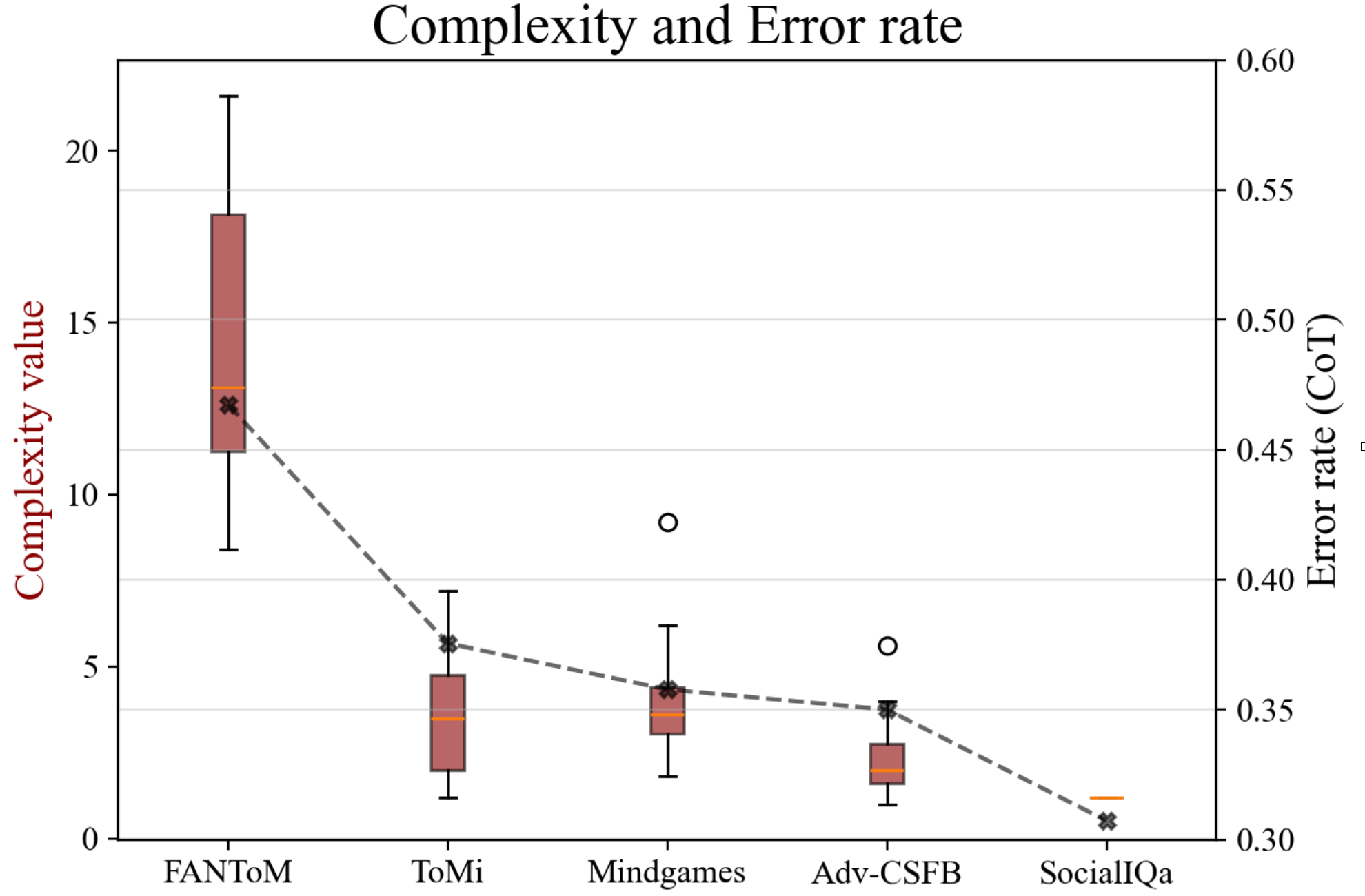} 
\caption{Each boxplot summarizes the complexity analysis of the five ToM benchmarks in ascending order. We report the average \textbf{error rate} (i.e., $1$-accuracy) of GPT-3.5-Turbo, GPT-4, Mixtral 8x7B and LLaMA3-70B on the task when prompted with CoT.
}
\label{fig:complexity-vs-accuracy} 
\end{figure}
\begin{figure*}
\includegraphics[width=1\textwidth]{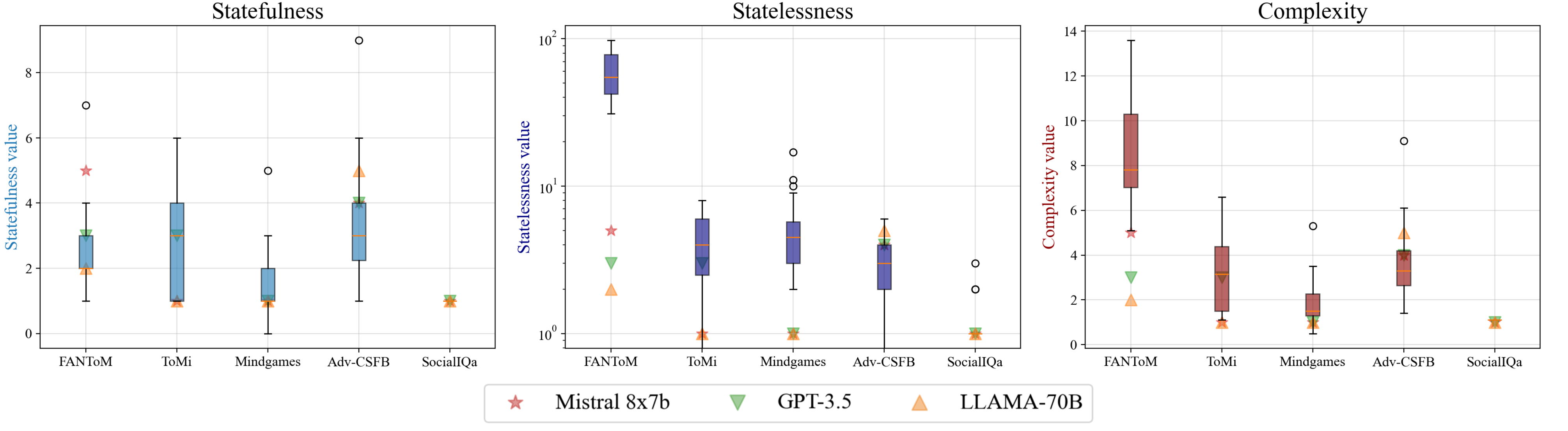} 
\caption{Each boxplot summarizes the statefulness (left), statelessness (middle, y-axis in log-scale) and complexity analysis (right) of the five ToM benchmarks. We report mean, standard deviation and outliers alongside the best DWM method (by the number of prompt splits) and observe a strong correlation between the number of splits and the statefulness.}
\label{fig:complexity-analysis} 
\end{figure*}
Recent works expressed concern about ToM benchmarks' efficacy in memorisation~\cite{jacovi2023stop,lamalfaCodeSimulationChallenges2024}.
This motivated us to quantify and then analyse the impact of memorisation of ToM benchmarks on performance.
We computed the percentage of memorised prompts to understand whether that affects the performance of techniques, such as DWM, that split the prompt into chunks and introduce additional information instead of CoT, which leaves the input prompt unchanged.
As illustrated in Table~\ref{tab:memorisation}, ToMi and FANToM have been heavily memorised, with entire portions of the benchmarks that can be retrieved \emph{word for word} from GPT-3.5-Instruct (the autocomplete model by OpenAI).
Despite that, no clear evidence of a performance drop in DWM induced by memorisation exists. For GPT-3.5, despite CoT having higher performance on ToMi, DWM is better on FANToM (Figure~\ref{fig:gpt-3.5-dwm-results}).
We hypothesise that as long as a memorised problem is prompted, either in its exact form (as for CoT) or split as in DWM, the most potent models can recover it alongside the ground truth label, thus invalidating the test for both.
We conclude with a note of caution: while we discovered that ToMi and FANToM are memorised by GPT-3.5-Instruct, that does not imply any LLM, including GPT-3.5-Turbo and GPT-4, whose training details are not released publicly, has been trained on that data.

\subsection{Statefulness of ToM Benchmarks}
We used the complexity framework introduced in Section~\ref{sec:complexity} to characterise the statefulness and statelessness of the five ToM benchmarks used for the experimental evaluation.
We randomly sampled $50$ problems from each dataset, identified the objects, and manually labelled stateful and stateless \emph{state events}. We release the split samples alongside a web application that facilitates manual labelling.
As illustrated in Figure~\ref{fig:complexity-analysis} (left), the statefulness of each problem, i.e., that of the object a model must track to answer correctly, \textbf{strongly correlates} with the best-performing DWM split. 
The statelessness complexity, reported in Figure~\ref{fig:complexity-analysis} (middle), i.e., that of objects that a model does not need to track, grows larger for problems such as FANToM, only partially influencing the models' performance. We hypothesise that the most potent models developed some competency in discerning the relevant part of a prompt (the stateful events) from the confounding ones. 
We finally report, in Figure~\ref{fig:complexity-analysis} (right), the complexity of each problem computed as per Eq.~\ref{def:tom-complexity}, with $\tau$ set in a range between $0.05$ and $0.2$ (i.e., the relative weight of stateless compared to stateful events). Results suggest that FANToM is the most difficult ToM task for humans and LLMs (see Figure~\ref{fig:gpt-3.5-dwm-results}), followed by ToMi (the second most difficult for LLMs as well) and Adv-CSFB (which seems easier than the others); in contrast, Mindgames and SocialIQa tend to be easier.
Finally, in Figure~\ref{fig:complexity-vs-accuracy}, we compare the accuracy of GPT-3.5-Turbo, GPT-4, Mixtral 8x7B and LLaMA3-70B when prompted with CoT (i.e., without split) on the five ToM benchmarks with the complexity of the task as per Def.~\ref{def:tom-complexity}. We observe a \textbf{strong correlation} between the error-rate and the complexity of a task, i.e., our framework correctly identifies the tasks that are harder both for humans and current state-of-the-art LLMs.

\section{Conclusions}
This paper introduces a complexity framework to measure the difficulty of Theory of Mind (ToM) problems. It quantifies the difficulty by tracking necessary states (stateful) and unnecessary states (stateless), with the latter discounted in the complexity computation. The framework evidences a strong correlation between complexity and model performance.
Inspired by this framework, we propose DWM, a prompting technique that splits a prompt into parts to query a model for a consistent representation of the environment and agents' beliefs. DWM outperforms CoT and ToT by extracting implicit but relevant information.



\section*{Limitations}

\paragraph{Higher order belief tracking.} Our theoretical framework reduces the problem of solving a belief ToM problem to finding the correct descriptions that need to be tracked. It extends seamlessly to tasks with much higher complexity, however, we have not had the opportunity to test this theory in those settings. We noticed that most theory of mind tasks available in the community only require one to five states to be correctly answered. A possible extension would be testing the theory upon tasks with higher state complexity, e.g. $k^{\text{th}}$-order belief tracking tasks. However, it is unclear whether this could be useful in real applications as most human belief tracking is limited to 5 or 6 orders~\cite{cargile1970note, dennettIntentionalStanceTheory1988}.

\paragraph{On task splitting methods.} It is not straightforward to automatically find the correct task splits in a manner that correctly describes the state. An LLM could find a way to split it by itself correctly and use those splits to answer the question. We attempted this approach, yet with a simple prompting method, the model splits every sentence, making the descriptions much noisier and less accurate. Future work could try to find the best splits automatically.

\paragraph{Memorization analysis.} Training and evaluating on the same dataset produces positively biased data on the model's performance. While running our benchmarks on ToMi, we discovered that the GPT-3.5 model had completely memorized parts of the dataset. This motivated us to extend the memorization test to the other tasks. We urge the research community to include a memorization section on every benchmark study with public datasets used in their works. This data is crucial to conduct fair and unbiased research on evaluating LLMs' abilities~\cite{jacovi2023stop}.
Future works will include an analysis of the memorisation rate of other ToM tasks alongside tests to quantify their impact on different models.

\paragraph{On element interactivity.}
Sweller~\cite{swellerElementInteractivityIntrinsic2010} proposes a measure of complexity for cognitive tasks that encompasses three main components, namely the \emph{intrinsic}, \emph{extraneous}, and \emph{germane} cognitive load. 
In its framework, which has wide applications in education, the \emph{intrinsic load} relates to the number of references, or interactions, between the elements of a problem, i.e., the information or concept that needs to be understood to answer the question. 
Our framework approximates the \emph{intrinsic} and \emph{extraneous} loads to be single sentences in a ToM problem, which is not assured to be the best measure.

\section*{Ethical Statement}
The datasets and pre-trained LLMs that we use are all publicly available. This paper focuses on ToM problems' hardness and prompting methods. We highlight that LLMs do not guarantee the production of factual data or correct reasoning steps, and the prompting methods developed here should not be regarded as \textit{the} source of truth in decision-making.


\bibliography{custom}

\clearpage 

\appendix

\section{Experimental Setup}
\subsection{Experimental Details}

Most of the language models used in this work follow the Language
Models as a Service (LMaaS) paradigm~\cite{lamalfaLanguageModelsService2023}. This model of service does not allow transparency and hinders reproducibility. Reproducibility is difficult to achieve as 
common software development frameworks, such as CI/CD pipeline, ease the update of the public service but change the underlying entity. From this, it follows that the model tested by the researcher could change
at any time. This is not solvable from the outside. Researchers have no control over the software engineering practices inside a LMaaS, but could set some parameters to offer the highest possible grade of reproducibility.
We set the temperature to zero or enable greedy decoding by default (this does not imply determinism even if model weights are not changed).~\footnote{The main explanation is the \url{https://github.com/pytorch/pytorch/issues/75240}{"non-deterministic cuda cores"} another could "be batched inference in sparse MoE models", see \url{https://152334h.github.io/blog/non-determinism-in-gpt-4/}{here}} In prompting methods where the creativity of the response is exploited for better performance, e.g., Tree of Thoughts~\cite{yaoTreeThoughtsDeliberate2023}, we set the temperature to $0.7$, the value proposed in the reference papers.

\paragraph{LMaaS providers.}

We use \href{https://huggingface.co/}{Huggingface} for Mixtral 8x7B.
Groq Cloud for LLama-3-8B and LLama-3-70B.
Microsoft sponsorship for GPT-3.5 and GPT-4 access.

\subsection{Prompting Templates}
We present the different prompting techniques, taking as an example the following prompt from ToMi and GPT-3.5-Turbo as the reference model:

\begin{lstlisting}
1. Benjamin entered the workshop.
2. Isabella entered the workshop.
3. Hannah entered the workshop.
4. Isabella hates the onion
5. Hannah hates the t-shirt
6. The pajamas is in the bottle.
7. Isabella moved the pajamas to the drawer.
8. Benjamin exited the workshop.
9. Isabella exited the workshop.
10. Benjamin entered the workshop.
\end{lstlisting}

\noindent And the following question:
\begin{lstlisting}
Where does Benjamin think that Isabella searches for the pajamas?
\end{lstlisting}


\noindent \textbf{Chain of Thought}
\begin{lstlisting}
Consider the following dialogue where multiple agents interact. At the end, I will ask you a question to answer.
Here's the dialogue:

1. Benjamin entered the workshop.
2. Isabella entered the workshop.
3. Hannah entered the workshop.
4. Isabella hates the onion
5. Hannah hates the t-shirt
6. The pajamas is in the bottle.
7. Isabella moved the pajamas to the drawer.
8. Benjamin exited the workshop.
9. Isabella exited the workshop.
10. Benjamin entered the workshop.

This is the end of the dialogue. Now, this is a question for you to answer.

Question: Where does Benjamin think that Isabella searches for the pajamas?

Think step by step, answer the question with one word and provide the answer between <answer></answer> tags.
For example, reply with <answer>vase</answer>.
\end{lstlisting}

\noindent \textbf{Tree of Thought}

\noindent We first prompt an LLM to propose different solution paths to solve a task.
\begin{lstlisting}
Consider the following dialogue where multiple agents interact. At the end, I will ask you a question to answer.
Here's the dialogue:

1. Benjamin entered the workshop.
2. Isabella entered the workshop.
3. Hannah entered the workshop.
4. Isabella hates the onion
5. Hannah hates the t-shirt
6. The pajamas is in the bottle.
7. Isabella moved the pajamas to the drawer.
8. Benjamin exited the workshop.
9. Isabella exited the workshop.
10. Benjamin entered the workshop.

Question: Where does Benjamin think that Isabella searches for the pajamas?

Think step by step and list all possible answers providing a single answer on each line.
\end{lstlisting}

\noindent We then pick the best idea via a majority vote over three agents simulated by the LLM itself:
\begin{lstlisting}[escapechar=\%]
Given a dialogue and several observation choices, decide which choice is most promising. Analyze each choice in detail, then conclude in the last line "The best choice is {{s}}", where s the integer id of the choice.
1. Benjamin entered the workshop.
2. Isabella entered the workshop.
3. Hannah entered the workshop.
4. Isabella hates the onion
5. Hannah hates the t-shirt
6. The pajamas is in the bottle.
7. Isabella moved the pajamas to the drawer.
8. Benjamin exited the workshop.
9. Isabella exited the workshop.
10. Benjamin entered the workshop.

Here are some possible observations:
%\textbf{\#\# Here we insert the output of the previous prompt.}% 
\end{lstlisting}

\noindent We eventually ask the model for a final answer. 
\begin{lstlisting}[escapechar=\%]
Given this dialogue and possible observations, answer the question with one word and provide the answer between <answer></answer> tags.
1. Benjamin entered the workshop.
2. Isabella entered the workshop.
3. Hannah entered the workshop.
4. Isabella hates the onion
5. Hannah hates the t-shirt
6. The pajamas is in the bottle.
7. Isabella moved the pajamas to the drawer.
8. Benjamin exited the workshop.
9. Isabella exited the workshop.
10. Benjamin entered the workshop.

Question: Where does Benjamin think that Isabella searches for the pajamas?

%\textbf{\#\# Here we insert the observations generated by the LLM with the previous prompts.}% 

For example, reply with <answer>vase</answer>.
\end{lstlisting}

\noindent \textbf{Discrete World Models - 1 Split}
\begin{lstlisting}
I give you a phrase of a dialogue between agents. I will reveal more parts of it later. At the end, I will give you a question you must answer. 
For each phrase, you must:
# 1. Write down a succinct description of what each agent knows about the environment and about the other agents. Keep the description short and do not produce redundant information. 
# 2. Each considerations you make must be preceded by the symbol #GPT#.
Here's the dialogue:

1. Benjamin entered the workshop.
2. Isabella entered the workshop.
3. Hannah entered the workshop.
4. Isabella hates the onion
5. Hannah hates the t-shirt
6. The pajamas is in the bottle.
7. Isabella moved the pajamas to the drawer.
8. Benjamin exited the workshop.
9. Isabella exited the workshop.
10. Benjamin entered the workshop.

This is the end of the dialogue. Now, this is a question for you to answer.

Question: Where does Benjamin think that Isabella searches for the pajamas?

Think step by step, answer the question with one word and provide the answer between <answer></answer> tags.
For example, reply with <answer>vase</answer>.
\end{lstlisting}

\noindent \textbf{Discrete World Model - 3 Split}
\begin{lstlisting}[escapechar=\%]
I give you a phrase of a dialogue between agents. I will reveal more parts of it later. At the end, I will give you a question you must answer. 
For each phrase, you must:
# 1. Write down a succinct description of what each agent knows about the environment and about the other agents. Keep the description short and do not produce redundant information. 
# 2. Each considerations you make must be preceded by the symbol #GPT#.
Here's the dialogue:

1. Benjamin entered the workshop.
2. Isabella entered the workshop.
3. Hannah entered the workshop.
%\textbf{\#\# Here the LLM provides a description of the environment so far described by the dialogue.}% 

4. Isabella hates the onion
5. Hannah hates the t-shirt
6. The pajamas is in the bottle.
%\textbf{\#\# Here the LLM provides a description of the environment so far described by the dialogue.}% 

7. Isabella moved the pajamas to the drawer.
8. Benjamin exited the workshop.
9. Isabella exited the workshop.
10. Benjamin entered the workshop.

This is the end of the dialogue. Now, this is a question for you to answer.

Question: Where does Benjamin think that Isabella searches for the pajamas?

Think step by step, answer the question with one word and provide the answer between <answer></answer> tags.
For example, reply with <answer>vase</answer>.
\end{lstlisting}

\noindent \textbf{Yaml/JSON}
\begin{lstlisting}[escapechar=\%]
Consider the following dialogue where multiple agents interact.

1. Benjamin entered the workshop.
2. Isabella entered the workshop.
3. Hannah entered the workshop.
4. Isabella hates the onion
5. Hannah hates the t-shirt
6. The pajamas is in the bottle.
7. Isabella moved the pajamas to the drawer.
8. Benjamin exited the workshop.
9. Isabella exited the workshop.
10. Benjamin entered the workshop.

Here is the YAML representation of the text.
%\textbf{\#\# Here we substitute the JSON/Yaml representation of the dialogue (see next prompt).}% 

Question: Question: Where does Benjamin think that Isabella searches for the pajamas?

Answer between the tags with a single word that is the answer of the above question
For example <answer>vase</answer>.
\end{lstlisting}

The JSON/YAML representation is required with the following prompt:
\begin{lstlisting}[escapechar=\%]
Consider the following dialogue where multiple agents interact.
1. Benjamin entered the workshop.
2. Isabella entered the workshop.
3. Hannah entered the workshop.
4. Isabella hates the onion
5. Hannah hates the t-shirt
6. The pajamas is in the bottle.
7. Isabella moved the pajamas to the drawer.
8. Benjamin exited the workshop.
9. Isabella exited the workshop.
10. Benjamin entered the workshop.

Now give a structured representation of the dialogue in YAML format. Keep track of the information that each agent has access to at each point in the dialogue.
It is important to have a relative representation of the information that each agent has access to at each point in the dialogue.
\end{lstlisting}

\section{Additional Results}
\subsection{DWM Prompting}\label{a:dwm-results}
\begin{figure*}
\centering 
\includegraphics[width=1\textwidth]{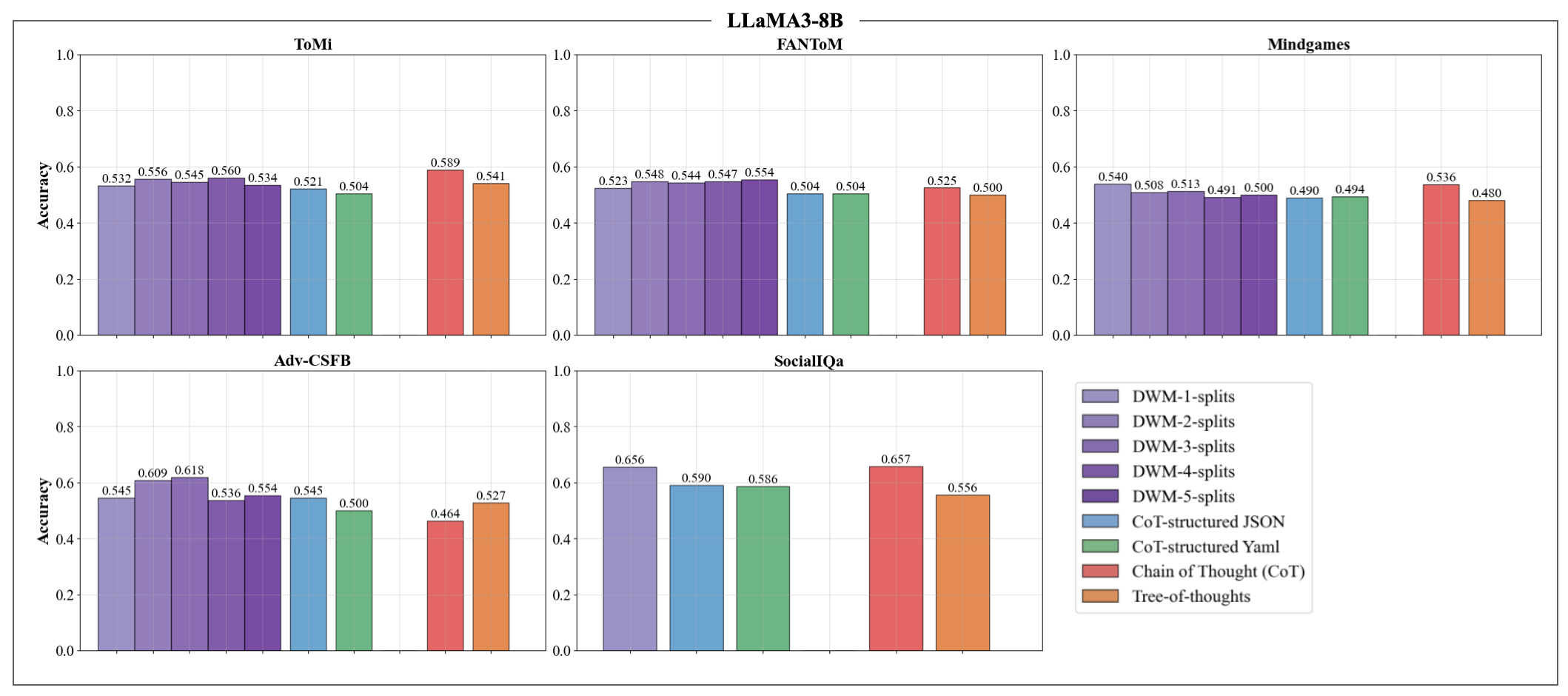} 
\includegraphics[width=1\textwidth]{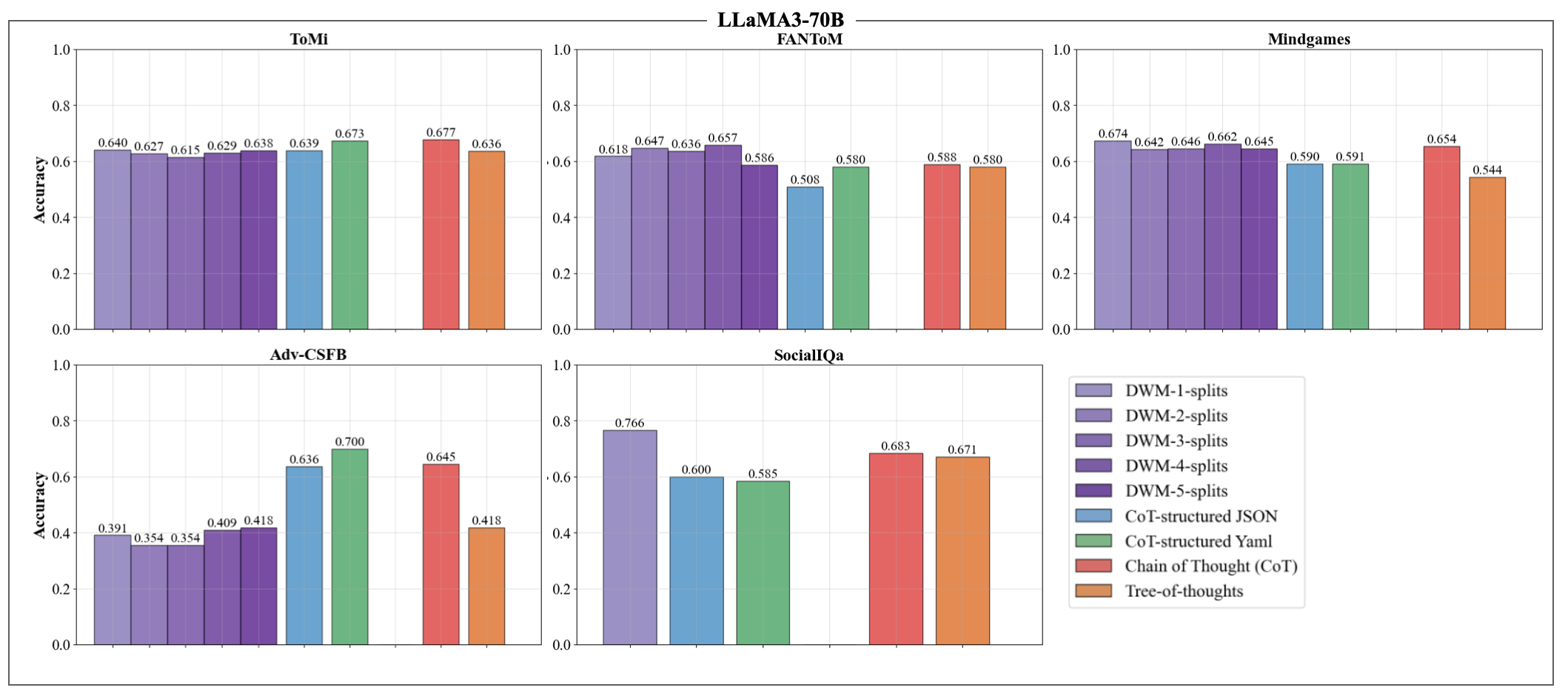} 
\includegraphics[width=1\textwidth]{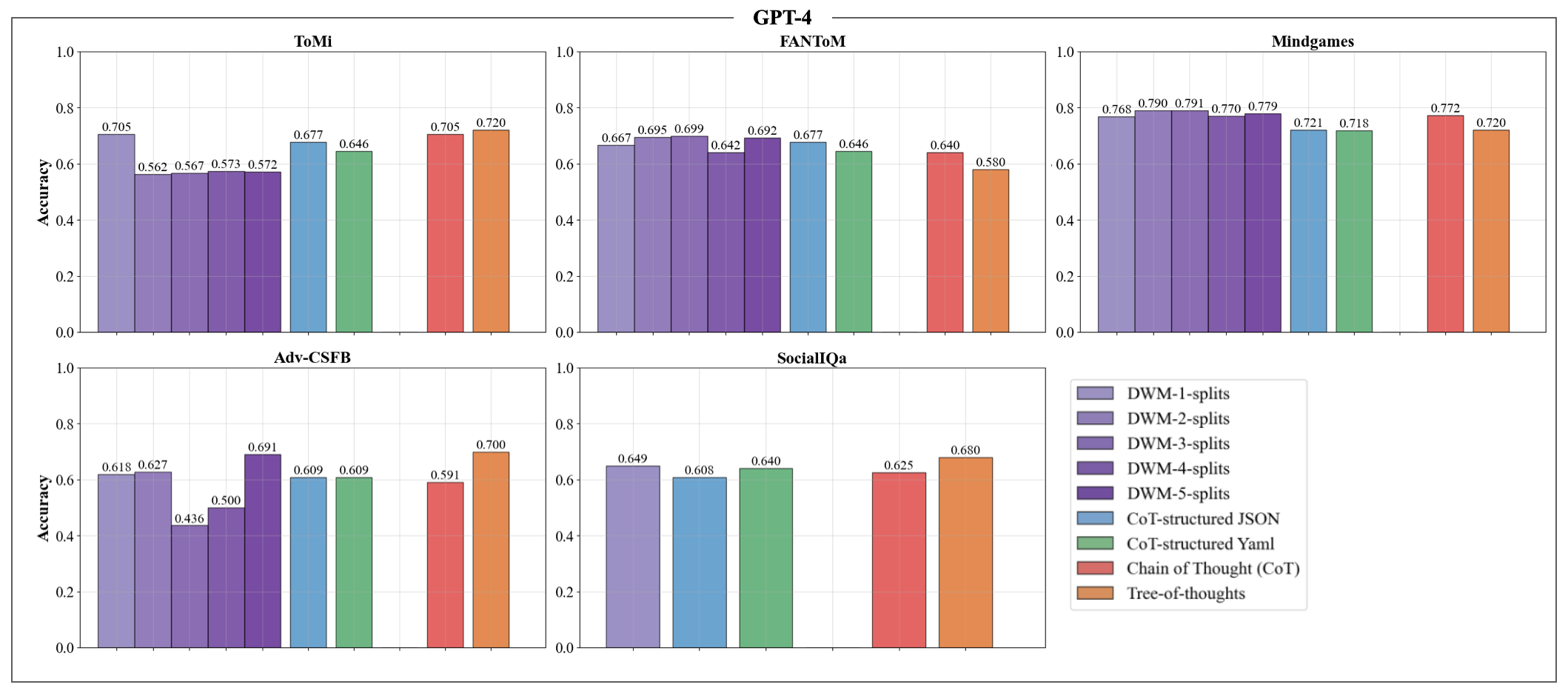} 
\caption{Benchmarks of LLaMA3-7B (top), LLaMA3-70B (middle) and GPT-4 (bottom) models on different ToM tasks for DWM (one to five splits), CoT, ToT and structured prompts (JSON and Yaml). For GPT-4 and ToT, we tested 50 samples (instead of 1000).}\label{fig:other-models-dwm-results}
\end{figure*}
In this section, and, in particular in Figure~\ref{fig:other-models-dwm-results}, we report results for LLaMA3-7B, LLaMA3-70B and GPT-4 on the five ToM benchmarks and for different prompting techniques, namely DWM (one to five splits), JSON, Yaml, CoT and ToT.

\subsection{DWM Elicits More Informed Mental States in LLMs}\label{a:dwm-vs-cot}
\begin{figure*}
\centering 
\includegraphics[width=1\textwidth]{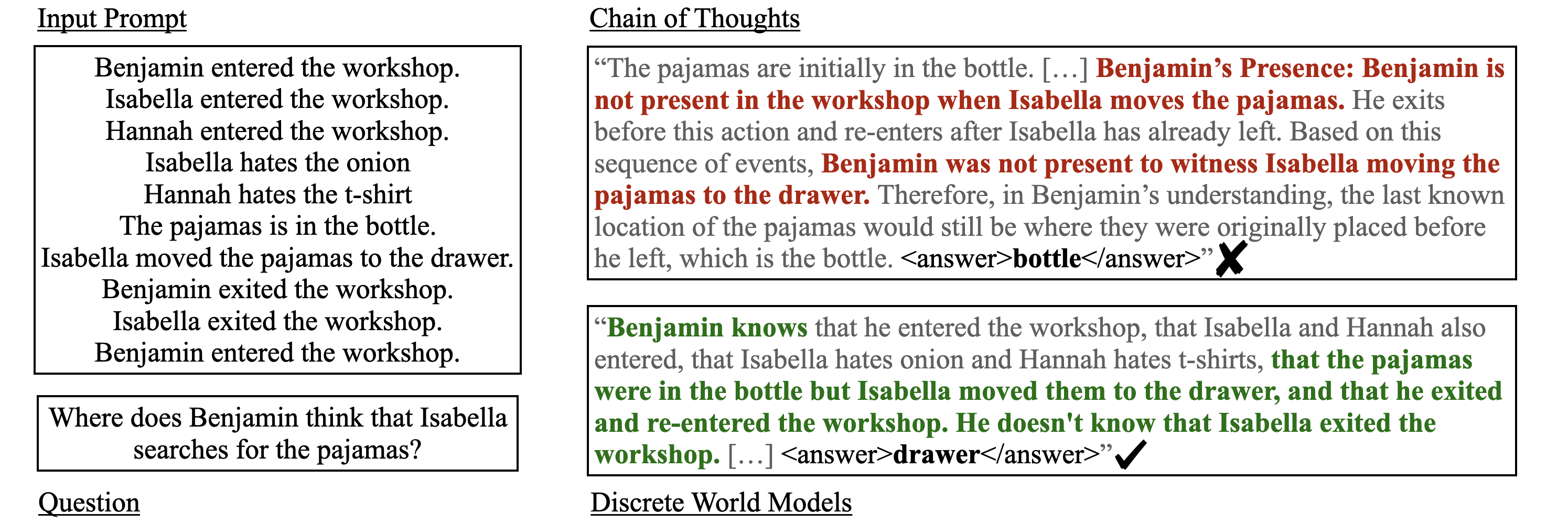} 
\caption{Example of a 
a ToMI instance
where GPT-4 fails when prompted with CoT, yet succeeds with DWM. CoT elicits an untruthful reasoning process (in \textbf{\textcolor{purple}{red}}), while DWM correctly informs the model with the correct information about Benjamin's first-order belief (in \textbf{\textcolor{ao(english)}{green}}).} 
\label{fig:a-dwm-explicit-tomi} 
\end{figure*}

\begin{figure*}
\centering 
\includegraphics[width=1\textwidth]{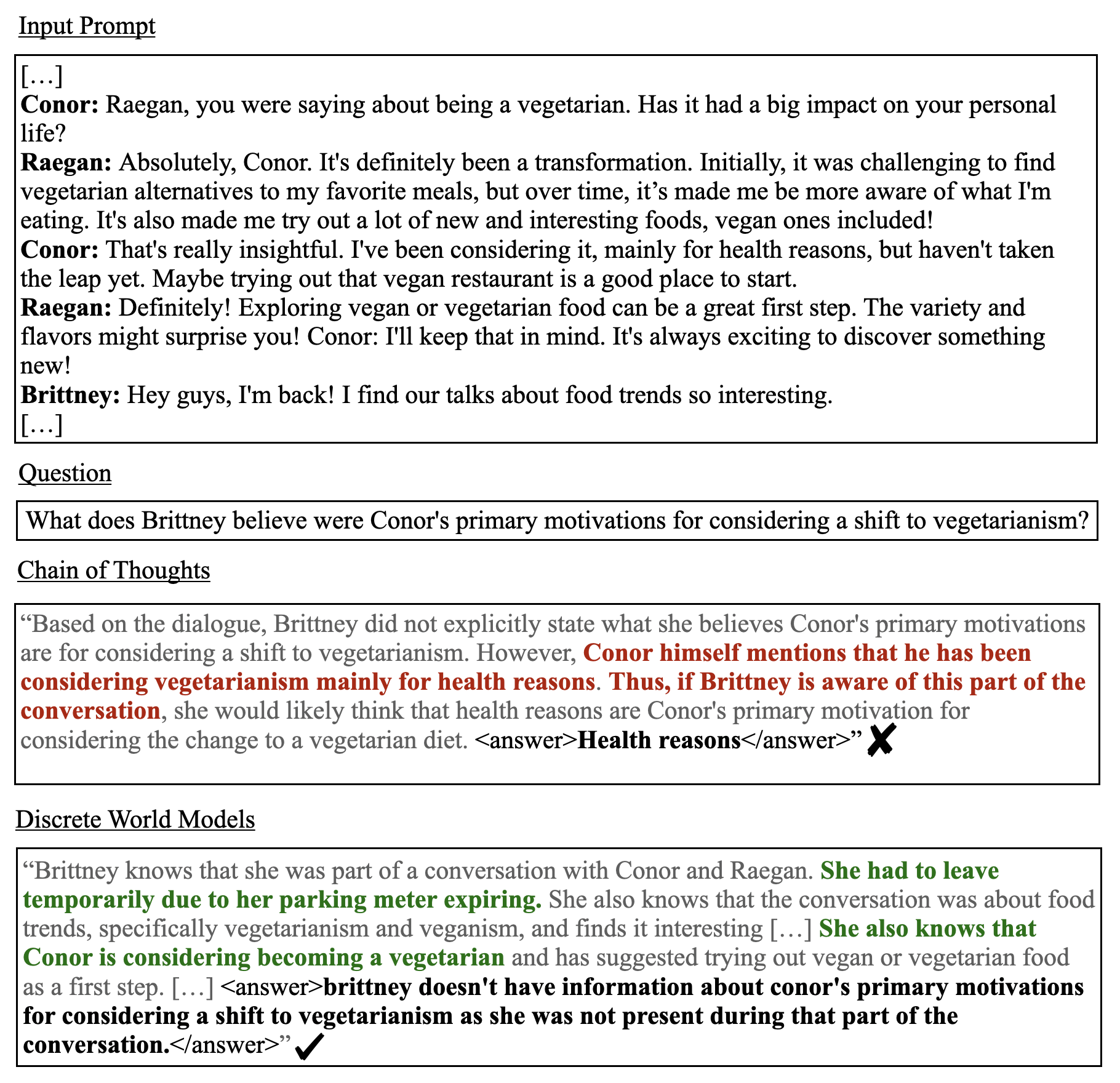} 
\caption{Example of a real FANToM example where GPT-4 fails when prompted with CoT, yet succeeds with DWM. CoT elicits an untruthful reasoning process (in \textbf{\textcolor{purple}{red}}), while DWM correctly informs the model with the correct information about the partial observability Brittney has about Conor (in \textbf{\textcolor{ao(english)}{green}}).} 
\label{fig:a-dwm-explicit-fantom} 
\end{figure*}

\begin{figure*}
\centering 
\includegraphics[width=1\textwidth]{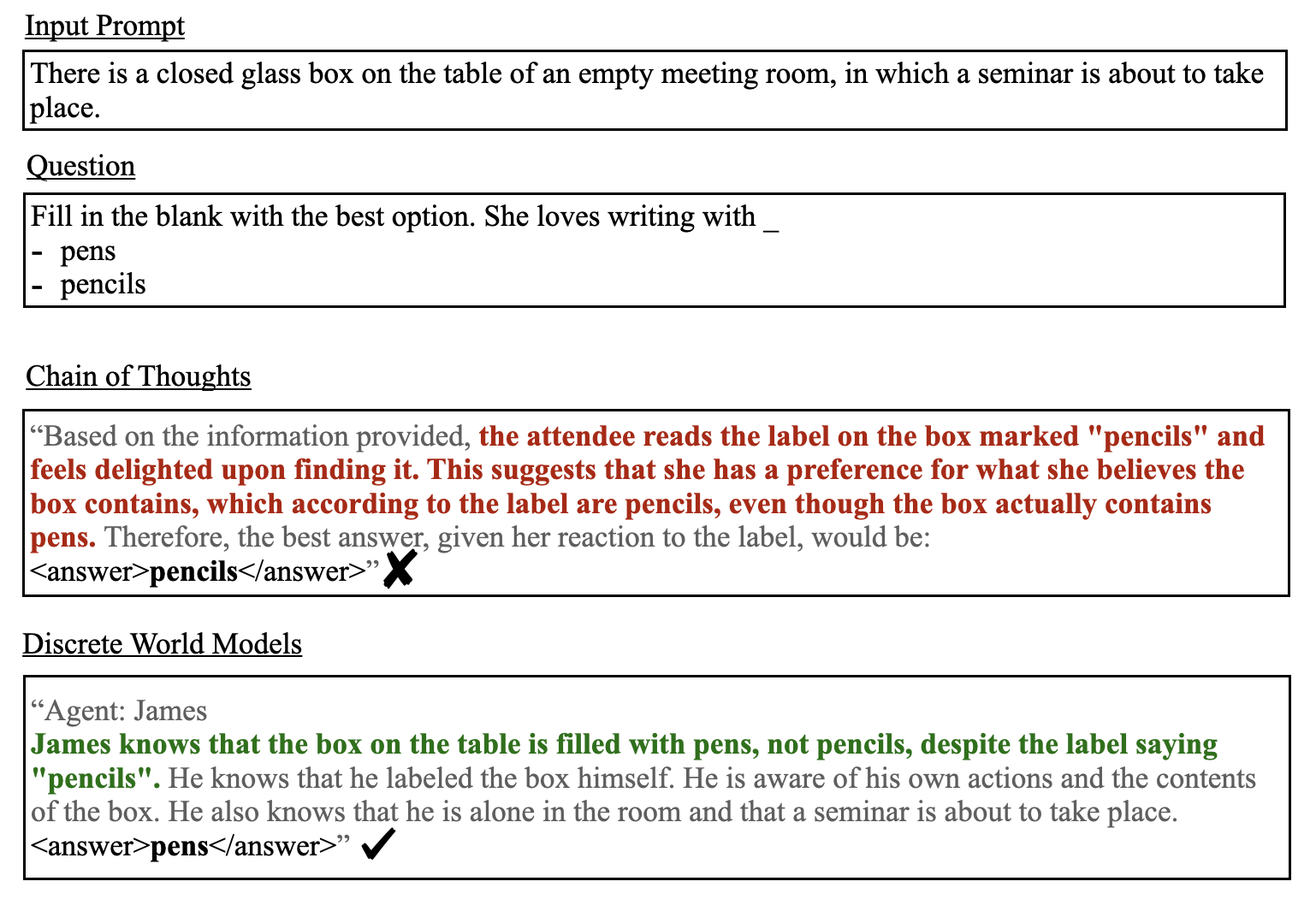} 
\caption{Example of a real ADV-csfb example where GPT-4 fails when prompted with CoT, yet succeeds with DWM. CoT elicits an untruthful reasoning process (in \textbf{\textcolor{purple}{red}}), while DWM correctly informs the model with the correct information about the content of the glass box (in \textbf{\textcolor{ao(english)}{green}}).} 
\label{fig:a-dwm-explicit-adv} 
\end{figure*}

\begin{figure*}
\centering 
\includegraphics[width=1\textwidth]{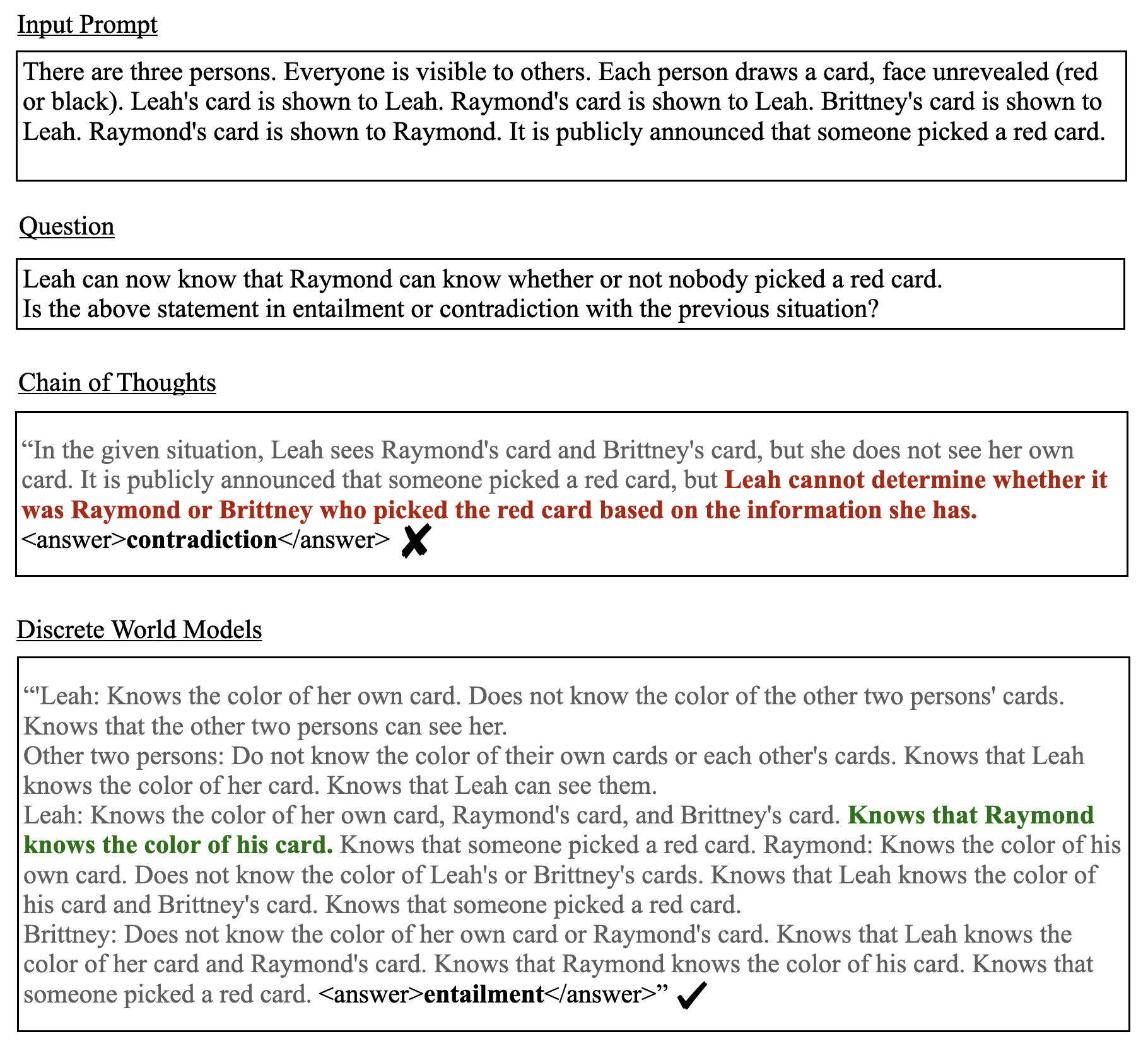} 
\caption{Example of a real Mindgames example where GPT-4 fails when prompted with CoT, yet succeeds with DWM. CoT elicits an untruthful reasoning process (in \textbf{\textcolor{purple}{red}}), while DWM correctly informs the model with the correct information about the knowledge Leah has about Raymond (in \textbf{\textcolor{ao(english)}{green}}).} 
\label{fig:a-dwm-explicit-mindgames} 
\end{figure*}

\begin{figure*}
\centering 
\includegraphics[width=1\textwidth]{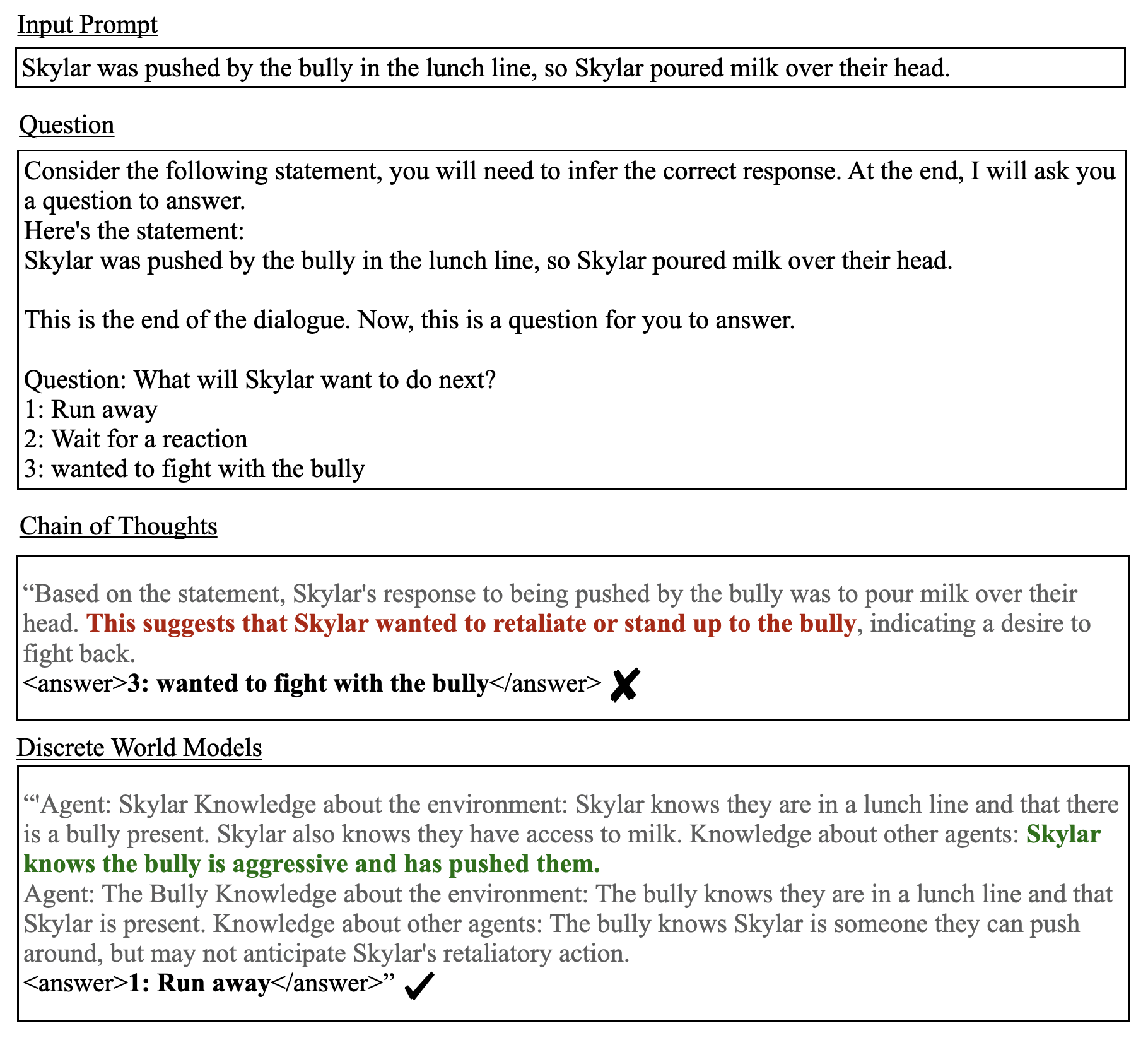} 
\caption{Example of a real SocialIQa example where GPT-4 fails when prompted with CoT, yet succeeds with DWM. CoT elicits an untruthful reasoning process (in \textbf{\textcolor{purple}{red}}), while DWM correctly informs the model with the correct next action Skylar will take (in \textbf{\textcolor{ao(english)}{green}}).} 
\label{fig:a-dwm-explicit-socialiqa} 
\end{figure*}

In this section, we report and discuss an example of a real prompt and the answers provided by GPT-4 for each ToM task we evaluated in this paper. For FANToM (Figure~\ref{fig:a-dwm-explicit-fantom}), we just reported the portion of the prompt that induces an unfaithful reasoning process in GPT-4, due to the prohibitive length of the input prompts. Results for ToMi, FANToM, ADV-csfb, Mindgames and SocialIQa are reported respectively in Figures~\ref{fig:a-dwm-explicit-tomi},~\ref{fig:a-dwm-explicit-fantom},~\ref{fig:a-dwm-explicit-adv},~\ref{fig:a-dwm-explicit-mindgames} and~\ref{fig:a-dwm-explicit-socialiqa}.

\end{document}